\documentstyle[latex-acl]{article}

\title{\vspace{-0.5in}{\small In Proceedings of the 32nd Annual
Meeting of the Association for Computational Linguistics,\\ pp 310-312, Association for Computational Linguistics, 1994.} \\ \LARGE\bf REAPING THE BENEFITS OF INTERACTIVE SYNTAX AND SEMANTICS\thanks{The author
would like to thank his advisor Dr. Kurt Eiselt and his colleague Justin
Peterson for their support and valuable comments on this work.}
}
\author{Kavi Mahesh\\
{\em Georgia Institute of Technology}\\
College of Computing \\
Atlanta, GA 30332-0280 USA \\
Internet: mahesh@cc.gatech.edu\\}

\input psfig

\begin{document}

\maketitle
\vspace{-0.5in}
\begin{abstract}
\begin{small}
Semantic feedback is an important source of information that a parser
could use to deal with local ambiguities in syntax. However, it is
difficult to devise a systematic communication mechanism for
interactive syntax and semantics. In this article, I propose a variant
of left-corner parsing to define the points at which syntax and
semantics should interact, an account of grammatical relations and
thematic roles to define the content of the communication, and a
conflict resolution strategy based on independent preferences from
syntax and semantics. The resulting interactive model has been
implemented in a program called COMPERE and shown to account for a
wide variety of psycholinguistic data on structural and lexical
ambiguities.
\end{small}
\end{abstract}

\section*{INTRODUCTION}
The focus of investigation in language processing research has moved
away from the issue of semantic feedback to syntactic processing
primarily due to the difficulty of getting the communication between
syntax and semantics to work in a clean and systematic way. However,
it is unquestionable that semantics does in fact provide useful
information which when fed back to syntax could help eliminate many an
alternative syntactic structure. In this article, I address three
issues in the communication mechanism between syntax and semantics and
provide a complete and promising solution to the problem of
interactive syntactic and semantic processing.

Since natural languages are replete with ambiguities at all levels, it
appears intuitively that a processor with incremental interaction
between the levels of syntax and semantics which makes the best and
immediate use of both syntactic and semantic information to eliminate
many alternatives would win over either a syntax-first or a
semantics-first mechanism.  In order to devise such an interactive
mechanism, one has to address three important issues in the
communication: (a) {\em When to communicate:} at what points should
syntax and semantics interact, (b) {\em What to communicate:} what and
how much information should they exchange, and (c) {\em How to agree:}
how to resolve any conflicting preferences between syntax and
semantics.

In this article, I propose (a) a particular variant of left-corner
parsing that I call {\it Head-Signaled Left Corner Parsing} (HSLC) to
define the points where syntax and semantics should interact, (b) an
account of grammatical relations based on thematic roles as a medium
for communication, and (c) a simple strategy based on syntactic and
semantic preferences for resolving conflicts in the communication.
These solutions were motivated from an analysis of a large body of
psycholinguistic data and account for a greater variety of
experimental observations on how humans deal with structural and
lexical ambiguities than previous models (Eiselt et al, 1993).  While
it also appears that the proposed interaction with semantics could
make improvements to the efficiency of the parser in dealing with real
texts, such a conclusion can only be drawn after an empirical
evaluation.

\section*{WHEN TO COMMUNICATE} 
Syntax and semantics should interact only at those times when one can
provide some information to the other to help reduce the number of
choices being considered. Only when the parser has analyzed a unit
that carries some part of the meaning of the sentence (such as a
content word) can semantics provide useful feedback perhaps using
selectional preferences for fillers of thematic roles.  We need to
design a parsing strategy that communicates with semantics precisely
at such points.  While pure bottom-up parsing turns out to be too
circumspect for this purpose, pure top-down parsing is too eager since
it makes its commitments too early for semantics to have a say.  A
combination strategy called Left Corner (LC) parsing is a good middle
ground making expectations for required constituents from the leftmost
unit of a phrase but waiting to see the left corner before committing
to a bigger syntactic unit (E.g., Abney and Johnson, 1991).  In LC
parsing, the leftmost child (the left corner) of a phrase is analyzed
bottom-up, the phrase is projected upward from the leftmost child, and
other children of the phrase are projected top-down from the phrase.

While LC parsing defines when to project top-down, it does not tell us
when to make attachments. That is, it does not tell when to attempt to
attach the phrase projected from its left corner to higher-level
syntactic units. Should it be done immediately after the phrase has
been formed from its left corner, or after the phrase is complete with
all its children (both required and optional adjuncts), or at some
intermediate point?  Since ambiguities arise in making attachments and
since semantics could help resolve such ambiguities, the points at
which semantics can help, determine when the parser should attempt to
make such attachments.

LC parsing defines a range of parsing strategies in the spectrum of
parsing algorithms along the ``eagerness'' dimension (Abney and
Johnson, 1991). The two ends of this dimension are pure bottom-up
(most circumspect) and pure top-down (most eager) parsers.  Different
LC parsers result from the choice of arc enumeration strategies
employed in enumerating the nodes in a parse tree.  In Arc Eager LC
(AELC) Parsing, a node in the parse tree is linked to its parent
without waiting to see all its children.  Arc Standard LC (ASLC)
Parsing, on the other hand, waits for all the children before making
attachments. While this distinction vanishes for pure bottom-up or
top-down parsing, it makes a big difference for LC Parsing.

In this work, I propose an intermediate point in the LC Parsing
spectrum between ASLC and AELC strategies and argue that the proposed
point, that I call Head-Signaled LC Parsing (HSLC), turns out to be
the optimal strategy for interaction with semantics. In this strategy,
a node is linked to its parent as soon as all the required children of
the node are analyzed, without waiting for other optional children to
the right. The required units are predefined syntactically for each
phrase; they are not necessarily the same as the `head' of the phrase.
(E.g., N is the required unit for NP, V for VP, and {\em NP for PP.})
HSLC makes the parser wait for required units before interacting with
semantics but does not wait for optional adjuncts (such as PP adjuncts
to NPs or VPs). The parsing spectrum now appears thus: \\
\noindent{\tt (Bottom-Up $\rightarrow$ Head-Driven $\rightarrow$  ASLC $\rightarrow$ HSLC $\rightarrow$ AELC $\rightarrow$ Top-Down)} \\

\begin{small}
\begin{tabbing}
{\bf Algorithm HSLC:} \\
aaa\=\kill
Given a grammar and an empty set as the initial \\
\> forest of parse trees, \\
For each word, \\
Add a new node $T_{w}$ to the current forest of \\
\> trees \{$T_{i}$\} for  each category for the \\
\> word in the lexicon \\
\> mark $T_{w}$ as a complete subtree \\
Repeat until there are no more complete trees \\
that can be attached to other trees, \\
\> Pro\=pose attachments for a complete \\
\> subtree $T_{j}$ \\
\> \> to a $T_{i}$ that is expecting $T_{j}$, or \\
\> \> to a $T_{i}$ as an optional constituent, or \\
\> \> to \= a new $T_{k}$ to be created if $T_{j}$ can be \\
\> \> \> the left corner (leftmost child) of $T_{k}$ \\
\> Select an attachment (see below) and attach \\
\> If a new $T_{k}$ was created, add it to the forest,\\
\> \> and make expectations for required units\\
\> \> of $T_{k}$ \\
\> If a $T_{i}$ in the forest has seen all its required \\
\> units, \\
\> \> Mark the $T_{i}$ as a complete subtree. 
\end{tabbing}
\end{small}

Consider a PP attachment ambiguity and the tree traversal labelings
produced by different LC parsers shown in Figure 1.  It can be seen
from Figure 1a that AELC attempts to attach the PP to the VP or NP
even before the noun in the PP has been seen. At this time, semantics
cannot provide useful feedback since it has no information on the role
filler for a thematic role to evaluate it against known selectional
preferences for that role filler. Thus AELC is too eager for
interactive semantics.  ASLC, on the other hand, does not attempt to
attach the VP to the S until the very end (Fig 1b). Thus even the
thematic role of the subject NP remains unresolved until the very end.
ASLC is too circumspect for interactive semantics. HSLC on the other
hand, attempts to make attachments at the right time for interaction
with semantics (Fig 1c).

\begin{figure}[htbp]
\begin{center}
\ \psfig{figure=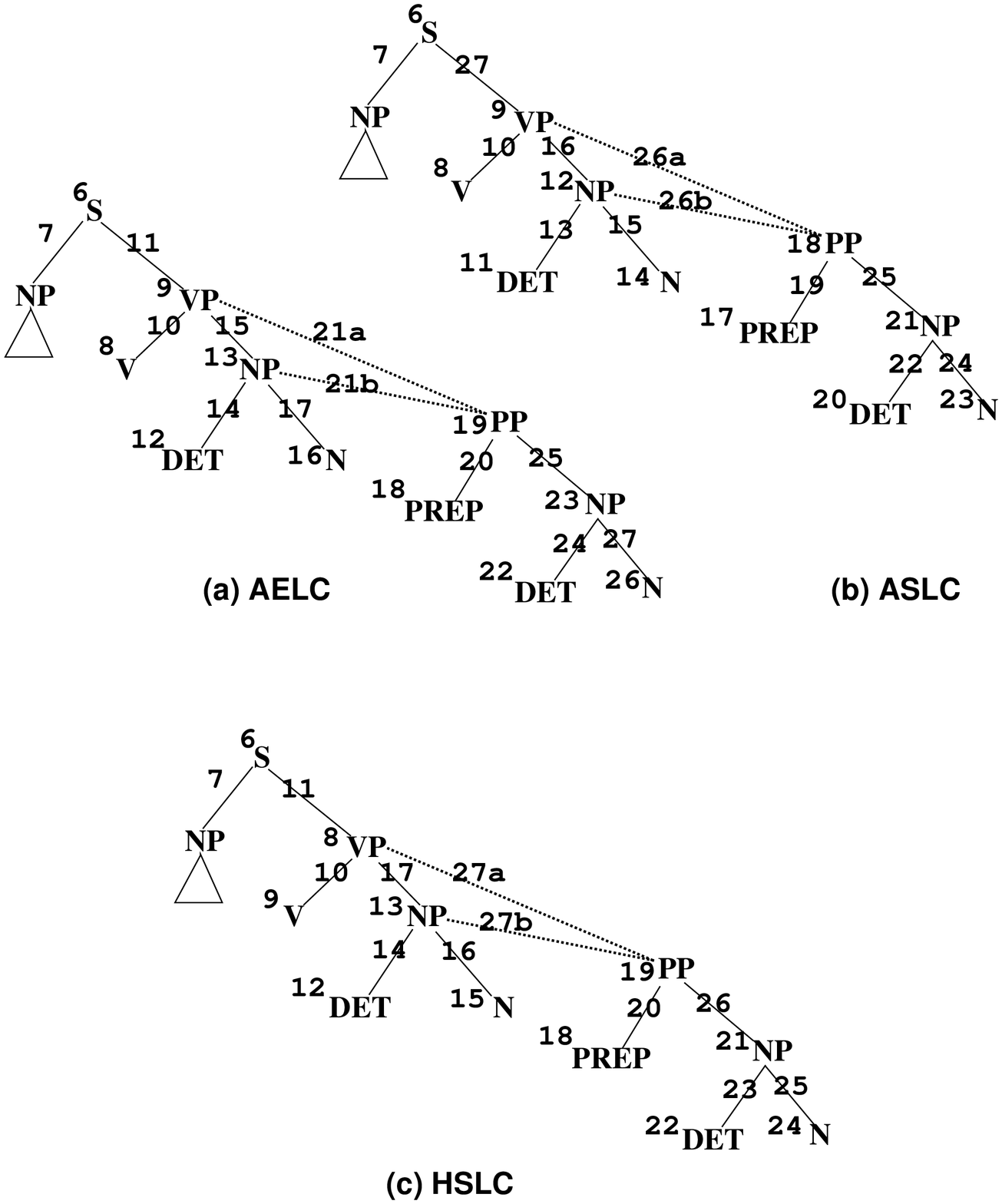,width=3.0in,height=3.1in}
\caption{LC Parsers at an Attachment Ambiguity}
\end{center}
\vskip -0.3in
\end{figure}

\section*{WHAT TO COMMUNICATE}
The content of the communication between syntax and semantics is a set
of grammatical relations and thematic roles. Syntax talks about the
grammatical relations between the parts of a sentence such as Subject,
Direct-object, Indirect-object, prepositional modifier, and so on.
Semantics talks about the thematic relations between parts of the
sentence such as event, agent, theme, experiencer, beneficiary,
co-agent, and so on. These two closed classes of relations are
translated to one another by introducing what I call ``intermediate
roles'' to take into account other kinds of linguistic information
such as active/passive voice, VP- {\em vs.} NP-modification, and so
on. Examples of intermediate roles are: active-subject,
passive-subject, VP-With-modifier, subject-With-modifier, and so on.
While space limitations do not permit a more detailed description
here, the motivation for intermediate roles as declarative
representations for syntax-semantics communication has been described
in (Mahesh and Eiselt, 1994).

The grammatical relations proposed by syntax are translated to the
corresponding thematic relations using the intermediate roles.
Semantics evaluates the proposed role bindings using any selectional
preferences for role fillers associated with the meanings of the words
involved. It communicates back to syntax a set of either an Yes, a No,
or a Don't-Care for each proposed syntactic attachment. A Yes answer
is the result of satisfying one more selectional preferences for the
role binding; a No for failing to meet a selectional constraint; and a
Don't-Care when there are no known preferences for the particular role
assignment.

\section*{HOW TO AGREE}
Since syntax and semantics have independent preferences for
multiple ways of composing the different parts of a sentence, an
arbitrating process (that I call the Unified Process) manages the
communication and resolves any conflicts. This unified process helps
select the alternative that is best given the preferences of both
syntax and semantics.  In addition, since the decisions so made are
never guaranteed to be correct, the unified process is not
deterministic and has the capability of retaining unselected
alternatives and recovering from any errors detected at later times.
The details of such an error recovery mechanism are not presented here
but can be found in (Eiselt et al, 1993) for example.

Syntax has several levels of preferences for the attachments it
proposes based on the following criteria: Attachment (of a required
unit) to an expecting unit has the highest preference.  Attachment as
an optional constituent to an existing (completed) unit has the next
highest preference.  Attachment to a node to be newly created (to
start a new phrase) has the least amount of preference.  These
preferences are used to rank syntactic alternatives.

\noindent{\bf The algorithm for the unified process:} \\
\begin{small}
Given: A set of feasible attachments \{$A_{i}$\} 
where each $A_{i}$ is a list of the two 
syntactic nodes being attached, the level of syntactic preference, and one of (Yes, No, Don't-Care) as the
semantic feedback,
\begin{tabbing}
If \= the \= most preferred syntactic alternative has\\
\> \> an Yes or Don't-Care, select it \\
\> else if no other syntactic alternative has a Yes,\\
\> \>  then select the most preferred syntactic\\
\> \> alternative that has a Don't-Care \\
\> else delay the decision and pursue multiple \\
\> \> interpretations in parallel until further \\
\> \> information changes the balance.
\end{tabbing}
\end{small}

\vskip -0.1 in

\section*{DISCUSSION}
The model of interactive syntactic and semantic processing proposed
accounts for a wide range psycholinguistic phenomena related to the
handling of lexical and structural ambiguities by human parsers. Its
theory of communication and the arbitration mechanism can explain data
that modular theories of syntax and semantics can explain as well as
data that interactive theories can (Eiselt et al, 1993). For instance,
it can explain why sentence (1) below is a garden-path but sentence
(2) is not.

\noindent{\bf (1)} The officers taught at the academy were very demanding. \\
\noindent{\bf (2)} The courses taught at the academy were very demanding. 

HSLC is different from both head-driven parsing and head-corner
parsing. It can be shown that the sequence of attachments proposed by
HSLC is more optimal for interactive semantics than those produced by
either of the above strategies. HSLC is a hybrid of left-corner and
head-driven parsing strategies and exploits the advantages of both.

In conclusion, I have sketched briefly a solution to the three
problems of synchronization, content, and conflict resolution in
interactive syntax and semantics. This solution has been shown to have
distinct advantages in explaining psychological data on human language
processing. The model is also a promising strategy for improving the
efficiency of syntactic analysis. However, the latter claim is yet to
be evaluated empirically.

\section*{REFERENCES}

\noindent{}Steven P. Abney and Mark Johnson. 1991. Memory Requirements and 
Local Ambiguities of Parsing 
Strategies. {\em J. Psycholinguistic Research,} 20(3):233-250.

\vskip 0.04 in

\noindent{}Kurt P. Eiselt, Kavi Mahesh, and Jennifer K. Holbrook. 1993. 
Having Your Cake and Eating It Too: Autonomy
and Interaction in a Model of Sentence Processing. {\em Proc.
Eleventh National Conference on Artificial Intelligence} (AAAI-93), pp 380-385.

\vskip 0.04 in

\noindent{}Kavi Mahesh and Kurt P. Eiselt. 1994. Uniform Representations
for Syntax-Semantics Arbitration. In  {\em Proc. Sixteenth
Annual Conference of the Cognitive Science Society,} Atlanta, GA, Aug 1994.

\end{document}